%% file: main.tex
\documentclass[letterpaper, 10 pt, conference]{ieeeconf}  
\IEEEoverridecommandlockouts                              
\usepackage{microtype}

\usepackage{graphics} 
\usepackage{epsfig} 
\usepackage{amsmath} 
\usepackage[sort,compress]{cite}%

\usepackage[pdftex,pdftitle={Safe Reinforcement Learning with Model Uncertainty Estimates}]{hyperref}
\hypersetup{colorlinks,linkcolor={green!50!black},citecolor={green!50!black},urlcolor={black}, pdfauthor={Bj{\"o}rn L{\"u}tjens, Michael F. Everett and Jonathan P. How}}
\makeatletter \let\NAT@parse\undefined \makeatother
\usepackage[sort,compress]{cite}
\usepackage{graphicx} 
\usepackage{amsfonts}
\usepackage{amsmath,soul}
\usepackage{color}
\usepackage[font=small]{subcaption}
\usepackage{balance}
\usepackage[font=small]{caption}
\usepackage[linesnumbered,ruled,vlined]{algorithm2e}
\usepackage{multirow}
\usepackage{dsfont}

\DeclareMathOperator*{\argmin}{\arg\!\min}
\DeclareMathOperator*{\argmax}{\arg\!\max}
\usepackage{tabulary}
\newcolumntype{K}[1]{>{\centering\arraybackslash}p{#1}}

\iffalse
\newif\ifPDF \ifx\pdfoutput\undefined\PDFfalse \else\ifnum\pdfoutput > 0\PDFtrue \else\PDFfalse \fi \fi

\ifPDF
\usepackage[pdftex, pdfstartview={FitV}, pdfpagelayout={TwoColumnLeft},bookmarksopen=true,plainpages = false, colorlinks=true, linkcolor=black, citecolor = black, urlcolor = black,filecolor=black, pagebackref=false,hypertexnames=false, plainpages=false, pdfpagelabels ]{hyperref}
\fi

\usepackage[capitalize]{cleveref}
\crefformat{equation}{(#2#1#3)}%
\Crefformat{equation}{Equations~(#2#1#3)}
\Crefname{equation}{Equation}{}

\makeatletter
\def\bstctlcite#1{\@bsphack
  \@for\@citeb:=#1\do{%
    \edef\@citeb{\expandafter\@firstofone\@citeb}%
    \if@filesw\immediate\write\@auxout{\string\citation{\@citeb}}\fi}%
  \@esphack}
\makeatother

\title{\LARGE \bf Safe Reinforcement Learning with Model Uncertainty Estimates 
}

\author{Bj{\"o}rn L{\"u}tjens, Michael Everett, Jonathan P. How
\thanks{\hspace*{-.17in} Aerospace Controls Laboratory, Massachusetts Institute of Technology, 77 Mass.\ Ave., Cambridge, MA, USA {\tt\footnotesize \{lutjens, mfe, jhow\}@mit.edu}}%
}

\usepackage[svgnames]{xcolor} \definecolor{DarkGreen}{rgb}{0,0.5,0}
\definecolor{DarkRed}{rgb}{0.75,0,0}

\usepackage[authormarkuptext=name,addedmarkup=bf,authormarkupposition=left]{changes}
\definechangesauthor[name={J.~H.}, color={blue}]{jh}
\definechangesauthor[name={M.~E.}, color={red}]{me}
\definechangesauthor[name={B.~L.}, color={green}]{bl}
\definechangesauthor[name={G.~H.}, color={orange}]{gh}

\usepackage{tikz,mathtools}
\usepackage[capitalize]{cleveref}
\crefformat{equation}{(#2#1#3)}
\Crefformat{equation}{Equation~(#2#1#3)}
\Crefname{equation}{Equation}{Equations}

\usetikzlibrary{shapes,positioning,automata,arrows,fit,backgrounds,calc}
\tikzstyle{block} = [draw, fill=blue!20, rectangle,minimum height=1em,
minimum width=2em] \tikzstyle{sum} = [draw, fill=blue!20, circle, node
distance=1cm] \tikzstyle{input} = [coordinate] \tikzstyle{output} =
[coordinate] \tikzstyle{pinstyle} = [pin edge={to-,thin,black}]
\usetikzlibrary{trees} \usetikzlibrary{decorations.pathmorphing}
\usetikzlibrary{decorations.markings}
\definecolor{darkgreen}{rgb}{0,0.5,0}
\definecolor{darkred}{rgb}{220,20,60}

\makeatletter
\renewcommand\paragraph{\@startsection{subsubsection}{4}{\z@}%
{0.25ex \@plus.5ex \@minus.2ex}%
{-.15em}%
{\normalfont\normalsize\itshape}}
\makeatother

\begin{document}


\maketitle
\thispagestyle{empty}
\pagestyle{empty}

\begin{abstract}
Many current autonomous systems are being designed with a strong reliance on black box predictions from deep neural networks (DNNs). However, DNNs tend to be overconfident in predictions on unseen data and can give unpredictable results for far-from-distribution test data. The importance of predictions that are robust to this distributional shift is evident for safety-critical applications, such as collision avoidance around pedestrians. Measures of model uncertainty can be used to identify unseen data, but the state-of-the-art extraction methods such as Bayesian neural networks are mostly intractable to compute. This paper uses MC-Dropout and Bootstrapping to give computationally tractable and parallelizable uncertainty estimates. The methods are embedded in a Safe Reinforcement Learning framework to form uncertainty-aware navigation around pedestrians. The result is a collision avoidance policy that \textit{knows what it does not know} and cautiously avoids pedestrians that exhibit unseen behavior. The policy is demonstrated in simulation to be more robust to novel observations and take safer actions than an uncertainty-unaware baseline.
\end{abstract}

\input{intro}

\input{related_work}
\input{approach}

\input{results}

\input{discussion}
\input{conclusion}





\section*{ACKNOWLEDGMENT}

This work is supported by Ford Motor Company. The authors want to thank Golnaz Habibi for insightful discussions.


\balance
\bibliographystyle{IEEEtran} 
\bibliography{biblio}

\end{document}

%% file: intro.tex

\section{Introduction} \label{sec:intro}

Reinforcement learning (RL) is used to produce state-of-the-art results in manipulation, motion planning and behavior prediction. However, the underlying neural networks often lack the capability to produce qualitative predictive uncertainty estimates and tend to be overconfident on out-of-distribution test data~\cite{Amodei_2016, Lakshmi_2016, Hendrycks_2017}. In safety-critical tasks, such as collision avoidance of cars or pedestrians, incorrect but confident predictions of unseen data can lead to fatal failure \cite{Tesla_2016}. We investigate methods for Safe RL that are robust to unseen observations and \textit{know what they do not know} to be able to raise an alarm in unpredictable test cases; ultimately leading to safer actions. 



\begin{figure}[t]
	\centering
    \includegraphics [trim=0 0 0 0, clip, width=1.0\columnwidth, angle = 0]{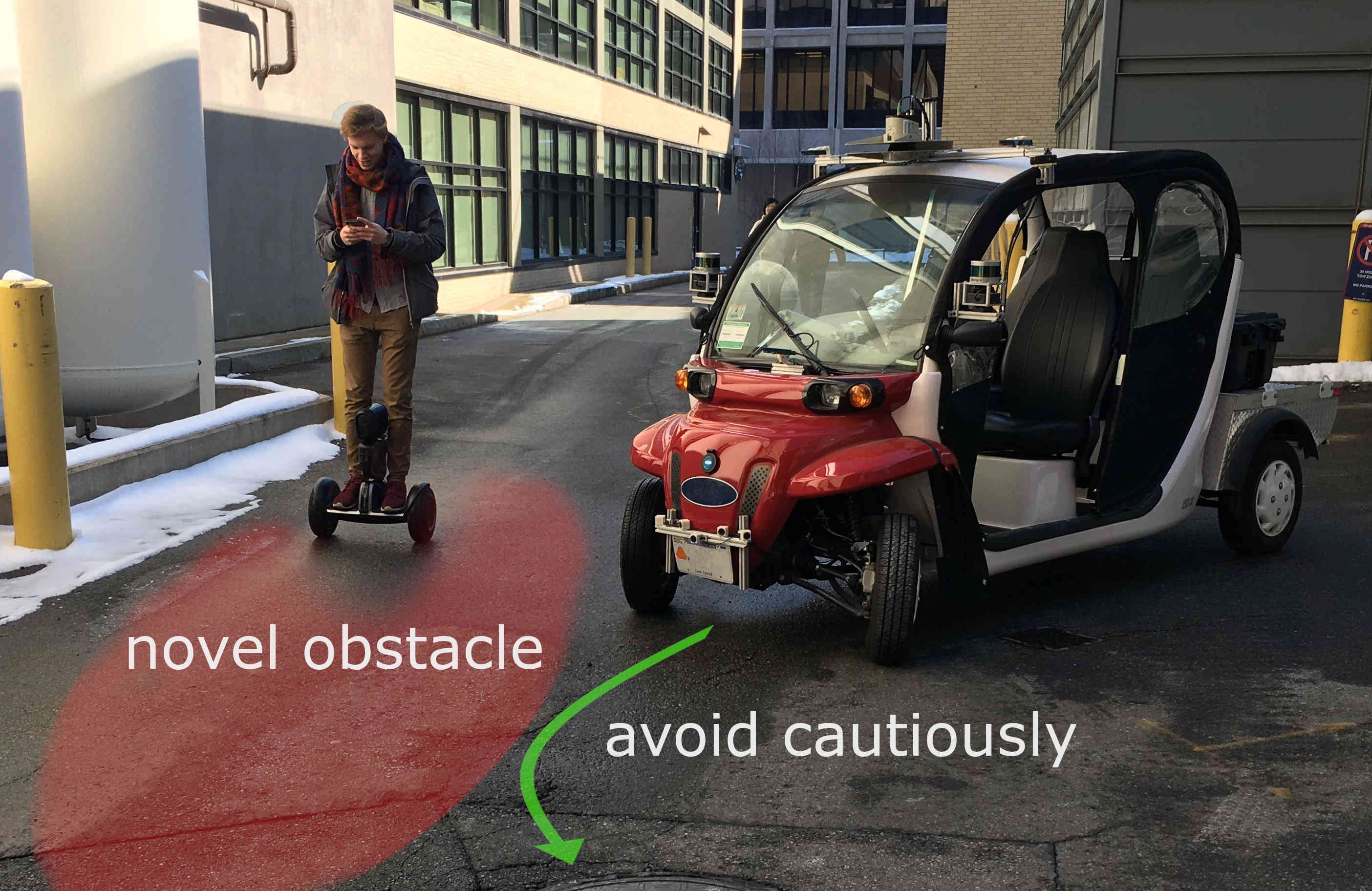}
    \label{fig:motivation} 
	\caption{An autonomous vehicle observes a novel dynamic obstacle that has never appeared during training, for example, an uncollaborative pedestrian on a personal vehicle. The proposed Reinforcement Learning framework detects the novelty and takes an action that cautiously avoids the pedestrian.}
  \vspace{-.2in}
\end{figure}

A particularly challenging safety-critical task is avoiding pedestrians in a campus environment with an autonomous shuttle bus or rover~\cite{Miller_2016,Navya_2018}. Humans achieve mostly collision-free navigation by understanding the hidden intentions of other pedestrians and vehicles and interacting with them~\cite{Zheng_2015, Helbing_1995}. Furthermore, most of the time this interaction is accomplished without verbal communication. Our prior work uses RL to capture the hidden intentions and achieve collaborative navigation around pedestrians~\cite{Chen_2016, Chen_2017, Everett_2018}. However, RL approaches always face the problem of generalizability from simulation to the real world and cannot guarantee performance on far-from-training test data. An example policy that has only been trained on collaborative pedestrians could fail to generalize to uncollaborative pedestrians in the real world, as seen in~\cref{fig:motivation}. The trained policy would output a best guess policy that might assume collaborative behavior and, without labeling the novel observation, fail ungracefully. To avoid such failure cases,  this paper develops a Safe RL framework for dynamic collision avoidance that expresses novel observations in the form of model uncertainty. The framework further reasons about the uncertainty and cautiously avoids regions of high uncertainty, as displayed in~\cref{fig:novelty_case}.


Much of the existing Safe RL research has focused on using external novelty detectors or internal modifications to identify environment or model uncertainty~\cite{Garcia_2015}. Note that our work targets  \textit{model uncertainty} estimates because they potentially reveal sections of the test data where training data was sparse and a model could fail to generalize~\cite{Gal_2016Thesis}. Work in risk-sensitive RL (RSRL) often focuses on \textit{environment uncertainty} to detect and avoid high-risk events that are known from training to have low probability but high cost~\cite{Geibel_2006, Mihatsch_2002, Shen_2013, Tamar_2015, Evendar_2006}. Other work in RSRL targets model uncertainty in MDPs, but does not readily apply to neural networks~\cite{Chow_2015, Mihatsch_2002}. Our work is mainly orthogonal to risk-sensitive RL approaches and could be combined into an RL policy that is robust to unseen data and sensitive to high-risk events.

Extracting model uncertainty from discriminatively trained neural networks is complex, as the model outcome for a given observation is deterministic. Mostly, Bayesian neural networks are used to extract model uncertainty but require a significant restructuring of the network architecture~\cite{Neal_1996}. Additionally, even approximate forms, such as Markov Chain Monte Carlo~\cite{Neal_1996} or variational methods~\cite{Blundell_2015, Graves_2011, Louizos_2016}, come with extensive computational cost and have a sample-dependent accuracy \cite{Neal_1996, Lakshmi_2016, Springenberg_2016}. Our work uses Monte Carlo Dropout (MC-Dropout)~\cite{Gal_2015} and bootstrapping~\cite{Osband_2016} to give parallelizable and computationally feasible uncertainty estimates of the neural network without significantly restructuring the network architecture~\cite{Dropout_2014, Bootstrap_1995}.

The main contributions of this work are i) an algorithm that identifies novel pedestrian observations and ii) avoids them more cautiously and safer than an uncertainty-unaware baseline, iii) an extension of an existing uncertainty-aware reinforcement learning framework \cite{Kahn_2017} to more complex dynamic environments with exploration aiding methods, and iv) a demonstration in a simulation environment.



%% file: related_work.tex

\section{Related Work} \label{sec:related_work}
This section investigates related work in Safe Reinforcement Learning to develop a dynamic collision avoidance policy that is robust to out-of-data observations.
\subsection{External verification and novelty detection}
Many related works use off-policy evaluation or external novelty detection to verify the learned RL policy~\cite{Richter_2017, Long_2018, Garcia_2015}. Reachability analysis could verify the policy by providing regional safety bounds, but the bounds would be too conservative in a collaborative pedestrian environment~\cite{Lygeros_1999, Majumdar_2016, Perkins_2003,Liebenwein_2018}. Novelty detection approaches place a threshold on the detector's novelty output and switch to a safety controller if the threshold is exceeded~\cite{Richter_2017}. However, switching to safety controllers is often abrupt and can generate uncomfortable, and unpredictable driving behavior. In our framework, the vehicle stays away from uncertain regions, as seen in~\cref{fig:out_of_dist}, to predictively avoid interventions by an underlying safety controller.

\subsection{Environment and model uncertainty}
This paper focuses on detecting novel observations via model uncertainty, also known as parametric or epistemic uncertainty \cite{Kendall_2017}. The orthogonal concept of environment uncertainty captures the uncertainty due to the imperfect nature of partial observations~\cite{Gal_2016Thesis}. For example, an observation of a pedestrian trajectory will, even with infinite training in the real-world, not fully capture the decision-making process of pedestrians and thus be occasionally ambiguous; will she turn left or right? The RL framework accounts for the unobservable decision ambiguity by learning a mean outcome~\cite{Gal_2016Thesis}. Model uncertainty, in comparison, captures how well a model fits all possible observations from the environment. It could be explained away with infinite observations and is typically high in applications with limited training data, or with test data that is far from the training data~\cite{Gal_2016Thesis}. Thus, the model uncertainty captures cases in which a model fails to generalize to novel test data and hints when one should not trust the network predictions~\cite{Gal_2016Thesis}.


\subsection{Measures of model uncertainty}
A new research topic adapts neural networks to express their model uncertainty~\cite{Blundell_2015, Gal_2015,Osband_2016}. Bootstrapping has been explored to generate approximate uncertainty measures to guide exploration \cite{Osband_2016}. By training an ensemble of networks on partially overlapping dataset samples they agree in areas of common data and disagree, and have a large sample variance, in regions of uncommon data \cite{Lakshmi_2016, Osband_2016}. Dropout can be interpreted similarly, if it is activated during test-time, and has been shown to approximate Bayesian inference in Gaussian processes~\cite{Dropout_2014, Gal_2015}. An alternative approach uses a Hypernet, a network that learns the weights of another network to directly give parameter uncertainty values, but was shown to be computationally very expensive~\cite{Pawlowski_2017}. An innovative, but controversial, approach retrieves Bayesian uncertainty estimates via batch normalization~\cite{Teye_2018}. This work uses MC-Dropout and bootstrapping to give computationally tractable uncertainty estimates.

\subsection{Applications of model uncertainty in RL}
Measures of model uncertainty have been used in RL very recently to speed up training by guiding the exploration into regions of high uncertainty \cite{Thompson_1933, Osband_2016, Liu_2017}. Kahn et al. used uncertainty estimates in model-based RL for static obstacle collision avoidance~\cite{Kahn_2017}. Instead of a model-based RL approach, one could argue to use model-free RL and draw the uncertainty of an optimal policy output $\pi^* = \argmax_{\pi}(Q)$. However, the uncertainty estimate would contain a mix from the uncertainties of multiple objectives and would not focus on the uncertain region of collision. Our work extends the model-based framework by~\cite{Kahn_2017} to the highly complex domain of pedestrian collision avoidance.~\cite{Kahn_2017} is further extended by using the uncertainty estimates for guided exploration to escape locally optimal policies, analyzing the regional increase of uncertainty in novel dynamic scenarios, using LSTMs, and acting goal-guided.


%% file: approach.tex

\section{Approach} \label{sec:approach}

\begin{figure}[t]
  \centering
    \includegraphics [trim=0 0 0 0, clip, angle=0, width=1.0\columnwidth,
  keepaspectratio]{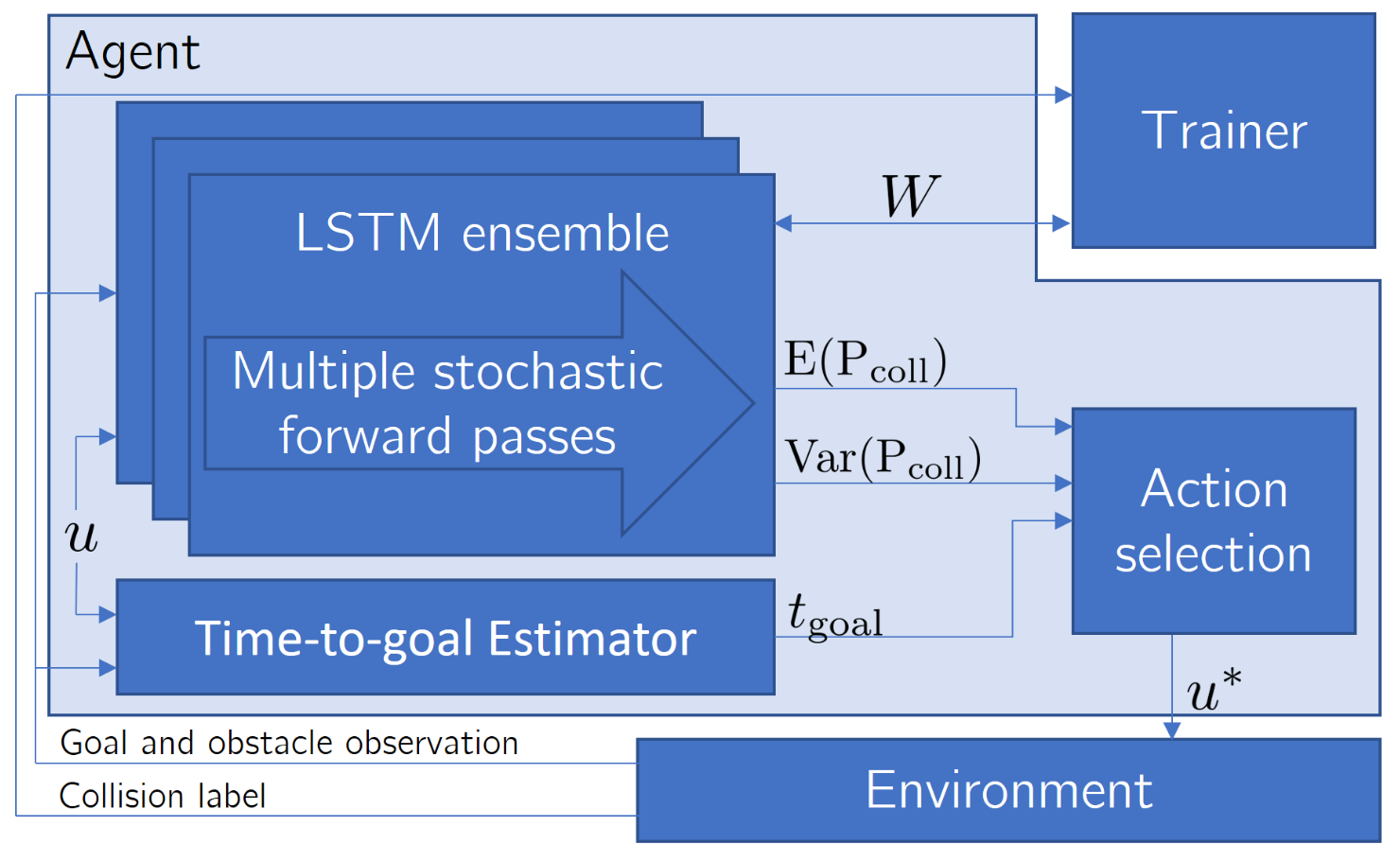}
  \caption{System architecture. An agent observes the environment and selects minimal cost motion primitives $u^*$ to reach a goal while avoiding collisions. On each time step, an ensemble of LSTM networks is sampled multiple times with different dropout masks to acquire a sample mean and variance collision probability for each motion primitive $u$.}
  \label{fig:sys_arch} 
    \centering
  \vspace{-.1in}
\end{figure}

This work proposes an algorithm that uses uncertainty information to cautiously avoid dynamic obstacles in novel scenarios. As displayed in the system architecture in~\cref{fig:sys_arch}, an agent observes a simulated obstacle's position and velocity, and the goal. A set of Long-Short-Term-Memory (LSTM)~\cite{Hochreiter_1997} networks predicts collision probabilities for a set of motion primitives $u$. MC-Dropout and bootstrapping are used to acquire a distribution over the predictions. From the predictions, a sample mean $\mathrm E(\mathrm P_{\text{coll}})$ and variance $\mathrm{Var}(\mathrm P_{\text{coll}})$ is drawn for each motion primitive. In parallel, a simple model estimates the time to goal $t_{\rm goal}$ at the end of each evaluated motion primitive. In the next stage, the minimal cost motion primitive $u^*$ is selected and executed for one step in the environment. The environment returns the next observation and at the end of an episode a collision label. After a set of episodes, the network weights $W$ are adapted and the training process continues. Each section of the algorithm is explained in detail below.


\subsection{Collision Prediction Network}
A set of LSTM networks (ensemble) predicts the collision probabilities of motion primitives. Each forward pass $i$ of a network returns the collision probability of an evaluated motion primitive: \begin{equation*}
\begin{aligned}
& \hspace*{.1in} \mathrm{P}^i_{\text{coll}} = \mathrm P^i\left(\mathds{1}_\text{coll} = 1\vert o_{t-l:t-1}, o_{t}, u_{t-l:t-1}, u_{t:t+h}\right)
\end{aligned}
\label{eq:p_coll} 
\end{equation*}
 where $\mathds{1}_\text{coll}$ is a collision label; $o_{t-l:t-1}$ is the history of observations in the last $l$ time steps; $o_t$ is the current observation; $u_{t-l:t-1}$ is a concatenation of past actions; and $u_{t:t+h}$ is the evaluated motion primitive of length $h$. The RL agent operates in a partially observable environment where it can only observe the pedestrian's position, velocity, and radius. The observation further contains the relative goal position of the RL agent. The motion primitive $u_{t:t+h}$ is element of a precomputed set of motion primitives $U$. In this work, $U$ contains 11 discrete motion primitives of length $h=1$ which are described by a heading angle $\alpha\in[-\frac{\pi}{6}, \frac{\pi}{6}]$. Regardless of the length, the optimal motion primitive is taken for one time step until the network is queried again. 

LSTM networks are chosen for the dynamic obstacle avoidance, because they are the state-of-the-art model in predicting pedestrian paths by understanding the hidden temporal intentions of pedestrians best~\cite{Alahi_2016_CVPR, Vemula_2017}. Based on this success, the proposed work first applies LSTMs to pedestrian avoidance in an RL setting. For safe avoidance, LSTM predictions need to be accurate from the first time step a pedestrian is observed in the robot's field of view. To handle the variable length observation input, masking~\cite{Che_2018} is used during training and test to deactivate LSTM cells that exceed the length of the observation history.


\subsection{Uncertainty Estimates with MC-Dropout and Bootstrapping}
MC-Dropout~\cite{Gal_2015} and bootstrapping~\cite{Osband_2016, Lakshmi_2016} are used to compute stochastic estimates of the model uncertainty $\rm Var(P_{coll})$. For bootstrapping, multiple networks are trained and stored in an ensemble. Each network is randomly initialized and trained on sample datasets that have been drawn with replacement from a bigger experience dataset~\cite{Osband_2016}. By being trained on different but overlapping sections of the observation space, the network predictions differ for uncommon observations and are similar for common observations. As each network can be trained and tested in parallel, bootstrapping does not come with significant computational cost and can be run on a real robot.

Dropout~\cite{Dropout_2014} is traditionally used for regularizing networks. It randomly deactivates network units in each forward pass by multiplying the unit weights with a dropout mask. The dropout mask is a set of Bernoulli random variables of value $[0,1]$, each with a keeping probability $p$. Traditionally, dropout is deactivated during test and each unit is multiplied with $p$. However,~\cite{Gal_2015} has shown that an activation of dropout during test, named MC-Dropout, gives model uncertainty estimates by approximating Bayesian inference in deep Gaussian processes. To retrieve the model uncertainty with dropout, our work executes multiple forward passes per network in the bootstrapped ensemble with different dropout masks ($p=0.7$) and acquires a distribution over predictions. For $n_d$ dropout samples in $n_b$ networks, a total of $N=n_d n_b$ forward passes are sampled. Although dropout has been seen to be overconfident on novel observations~\cite{Osband_2016},~\cref{tab:incr_uncertainty} shows that the combination of bootstrapping and dropout reliably detects novel scenarios.

From the parallelizable collision predictions from each network and each dropout mask, the sample mean and variance is drawn.

\subsection{Selecting actions}
A Model Predictive Controller (MPC) selects the safest motion primitive with the minimal joint cost:
\begin{equation*}
\begin{aligned}
& \hspace*{.05in} u^\star_{t:t+h} = \argmin_{u\in U}\left(\lambda_v {\mathrm{Var}_N}(\mathrm P^i_{\text{coll}}) + \right. \left. \lambda_c \mathrm E_N(\mathrm P^i_{\text{coll}}) + \lambda_g t_{\text{goal}}\right)
\end{aligned}
\label{eq:min_cost} 
\end{equation*}
The chosen MPC that considers the second order moment of probability~\cite{Lee_2017, Theodorou_2010, Kahn_2017} is able to select actions that are more certainly safe. The first and second order moment ($\mathrm E(\cdot)$ and $\mathrm Var(\cdot)$) are computed over the $N$ forward passes per motion primitive. The MPC estimates the time-to-goal $t_{\rm goal}$ from the end of each motion primitive by measuring the straight line distance. Each cost term is weighted by its own factor $\lambda$. Note that the soft constraint on collision avoidance requires $\lambda_g$ and $\lambda_c$ to be chosen such that the predicted collision cost $\lambda_c \mathrm E_N(\mathrm P^i_{\text{coll}}) ( \leq \lambda_c)$ is greater than the goal cost $\lambda_g t_{\text{goal}}$. In comparison to~\cite{Kahn_2017}, this work does not multiply the variance term with the selected velocity. The reason being is that simply stopping or reducing one's velocity is not always safe, for example on a highway scenario or in the presence of adversarial agents. The proposed work instead focuses on identifying and avoiding uncertain observations regionally in the ground plane.
 
\subsection{Adaptive variance}
Note that during training an overly uncertainty-averse model would discourage exploration and rarely find the optimal policy. Additionally, the averaging over multiple forward passes during prediction reduces the ensemble's diversity, which additionally hinders explorative actions. The proposed approach increases the penalty on highly uncertain actions $\lambda_v$ over time to overcome this effect. Thus, the policy efficiently explores in directions of high model uncertainty during early training phases; $\lambda_v$ is brought to convergence to act uncertainty-averse during execution. This work linearly increases $\lambda_v$ in $[-50000, 200]$ and has $\lambda_g = 2$, and $\lambda_c = 25$. 

\subsection{Collecting the dataset}
The selected action is executed in the learning environment. At the end of each episode $t_\text{end}$, the environment returns a collision label $\mathds{1}_{\text{coll}}$. The collision label is one if a collision occured during the episode and zero otherwise. The history of observations $o_{t_{start}:t_{\text{end}}}$ and actions $u_{t_{start}:t_{\text{end}}}$ from start to end of an episode is associated with the collision label and stored in an experience dataset. After running several episodes, random subsets from the full experience set are drawn to train the ensemble of networks for the next observe-act-train cycle. The policy roll-out cycle is necessary to learn how dynamic obstacles will react to the agent's learned policy. A supervised learning approach, as taken in~\cite{Richter_2017} for static obstacle avoidance, would not learn the reactions of environment agents on the trained policy.

%% file: results.tex
\section{Results} \label{sec:results}
We show that our algorithm uses uncertainty information to regionally detect novel obstacle observations and causes fewer collisions than an uncertainty-unaware baseline. First, a simple 1D case illustrates how the model regionally identifies novel obstacle observations. In a scaled up environment with novel multi-dimensional observations, the proposed model continues to exhibit regionally increased uncertainty values. The model is compared with an uncertainty-unaware baseline in a variety of novel scenarios; the proposed model performs more robust to novel data and causes fewer collisions.

\subsection{Regional novelty detection in 1D}
\begin{figure}[t]
  \centering
  \begin{subfigure}{1.0\columnwidth}
      \centering
      \includegraphics [trim=0 0 0 0, clip, width=1.0\textwidth, angle = 0]{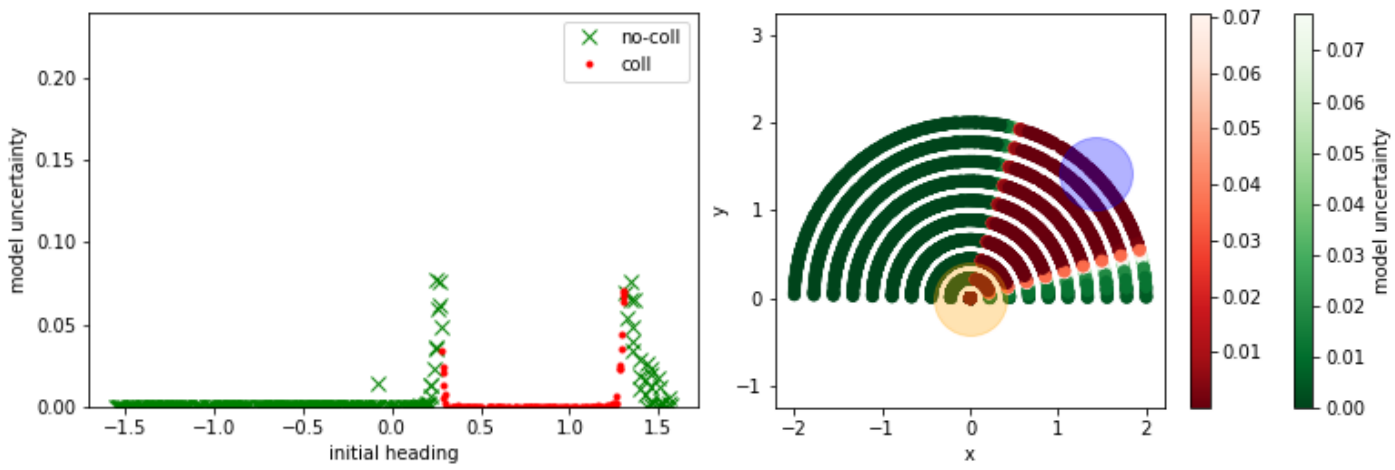}
      \caption{Known obstacle: low uncertainty}
      \label{fig:out_of_dist_train} 
      \vspace{+0.05in}
  \end{subfigure}
  \begin{subfigure}{1.0\columnwidth}
      \centering
      \includegraphics [trim=0 0 0 0, clip, width=1.0\textwidth, angle = 0]{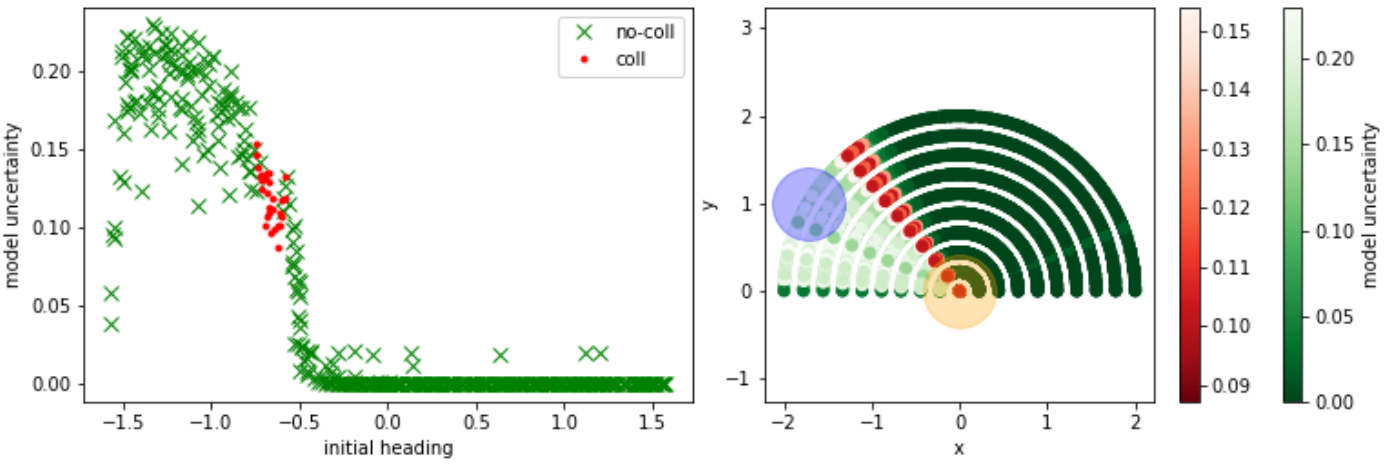}
      \caption{Novel obstacle: high uncertainty}
      \label{fig:out_of_dist_test} 
  \end{subfigure}
  \caption{Regional novelty detection in 1D. A simple network predicts collision (red) and no-collision (green) labels, given the agent's (orange) heading (left plot: x-axis) and a one-dimensional observation of an obstacle (blue) heading. The network accurately predicts labels with low uncertainty, when tested on the training dataset (a) . When tested on a novel observation set (b), the networks fails to predict accurate decision labels, but identifies them with a high regional uncertainty  (bottom-left: green points with high values, bottom-right: light green lines). Rather than believing in the false-positive collision predictions,~\Cref{fig:train_test} depicts how an agent would take a certainly safe action (dark green) to cautiously avoid the novel obstacle.}
  \label{fig:out_of_dist} 
  \vspace{+0.0in}
\end{figure}

First, we show that model uncertainty estimates are able to detect novel one-dimensional observations regionally, as seen in~\cref{fig:out_of_dist}. For the 1D test-case, a two-layer fully-connected network with MC-Dropout and Bootstrapping is trained to predict collision labels. To generate the dataset, an agent randomly chose heading actions, independent of the obstacle observations, and the environment reported the collision label. The network input is the agent heading angle and obstacle heading. Importantly, the training set only contains obstacles that are on the right-hand side of the agent (top plot:$x>0$).

\begin{figure}[t]
  \centering
  \begin{subfigure}{0.51\columnwidth}
      \centering
      \includegraphics [trim=0 0 0 0, clip, width=1.0\textwidth, angle = 0]{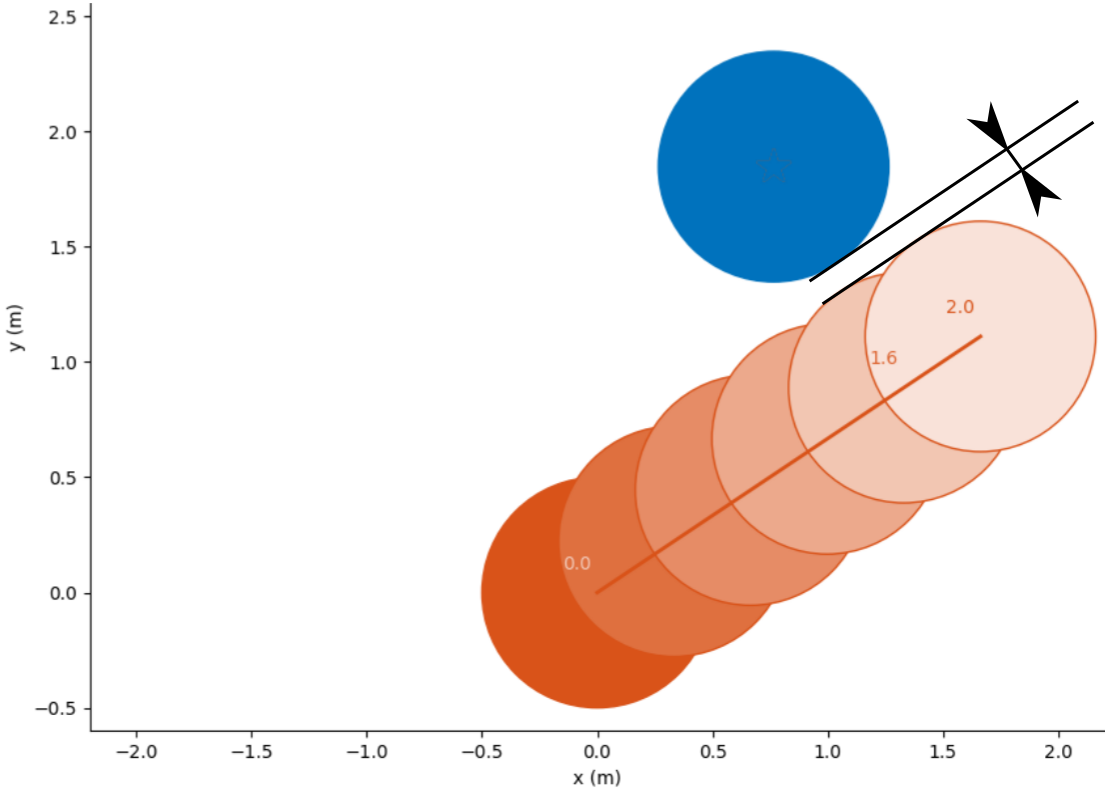}
      \caption{Known obstacle, confident}
      \label{fig:train} 
  \end{subfigure}
  \begin{subfigure}{0.455\columnwidth}
      \centering
      \includegraphics [trim=0 0 0 0, clip, width=1.0\textwidth, angle = 0]{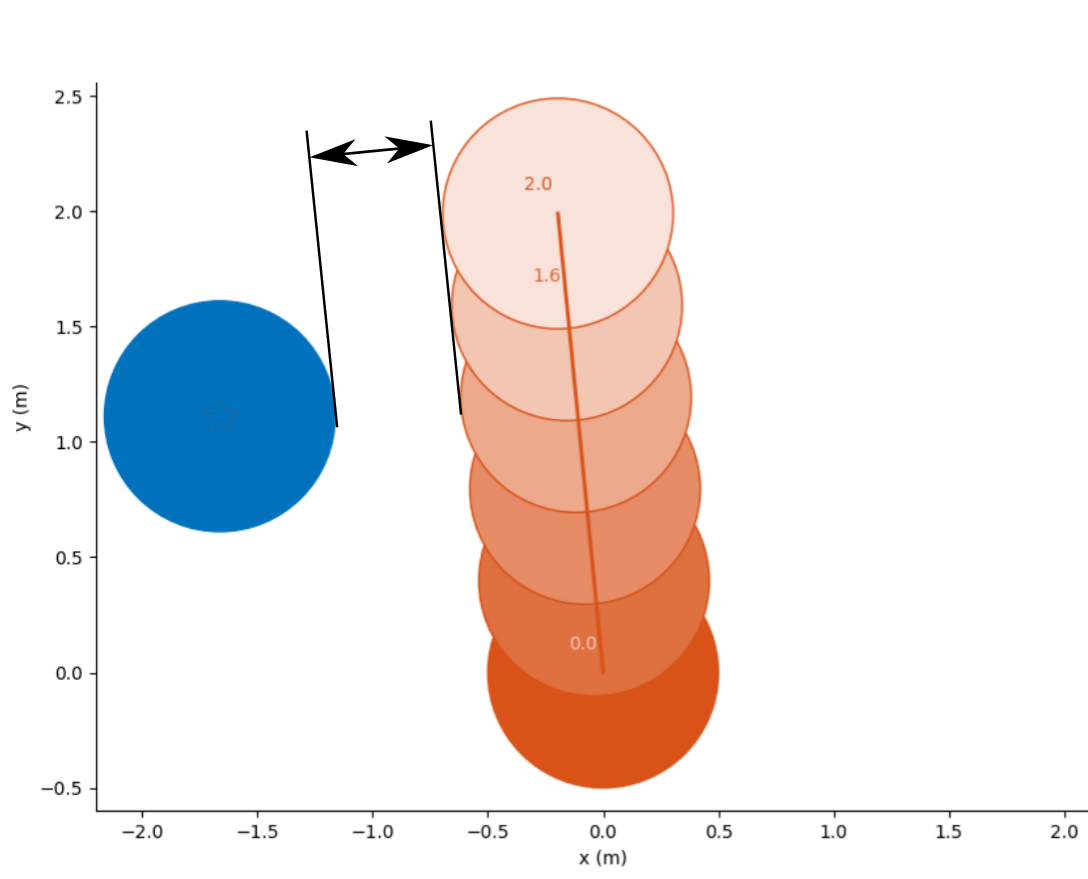}
      \caption{Novel obstacle, cautious}
      \label{fig:test} 
  \end{subfigure}
  \caption{Cautious avoidance after regional novelty detection. An agent (orange) in~\cref{fig:train} uses the uncertainty estimates from~\cref{fig:out_of_dist_train} to avoid a known obstacle (blue) confidently close. In~\cref{fig:test}, an agent recognizes a novel obstacle appearance, as seen in~\cref{fig:out_of_dist_test}, and cautiously avoids the obstacle.}
  \label{fig:train_test} 
  \vspace{-.2in}
\end{figure}

After training, the network accurately predicts collision and no-collision labels with low uncertainty for obstacle observations from the training distribution, as seen in~\cref{fig:out_of_dist_train}. For out-of-training obstacle observations on the agent's left (bottom plot: $x<0$), the neural network fails to generalize and predicts collision (red) as well as non-collision (green) labels for actions (straight lines) that would collide with the obstacle (blue). However, the agent identifies regions of high model uncertainty (left: y-axis, right: light colors) for actions in the direction of the unseen obstacle. The high uncertainty values suggest that the network predictions are false-positives and should not to be trusted. Based on the left-right difference in uncertainty estimates, the MPC would prefer a conservative action that is certainly safe (bottom-right: dark green lines) over a false-positive action that is predicted to be safe but uncertain (bottom-right: light green lines).~\Cref{fig:train_test} illustrates how the MPC chooses a conservative action to avoid a novel obstacle and confident actions to avoid known obstacles.

\subsection{Novelty detection in multi-dimensional observations}
The following experiments show that our model continues to regionally identify uncertainty in multi-dimensional observations and choose safer actions.
\subsubsection{Experiment setup}
A one-layer 16-unit LSTM model has been trained in a \textit{gym}~\cite{Gym_2016} based simulation environment with one agent and one dynamic obstacle. The dynamic obstacle in the environment is capable of following a collaborative RVO~\cite{Berg_2009}, GA3C-CADRL~\cite{Everett_2018}, or non-cooperative or static policy. For the analyzed scenarios, the agent was trained with obstacles that follow an RVO policy and are observed as described in~\cref{sec:approach}. The training process took 20 minutes on a low-compute amazon AWS c5.large Intel Xeon Platinum 8124M with 2vCPUs and 4GiB memory. Each of the used five networks in the ensemble is sampled twenty times by stochastic MC-Dropout forward passes. Drawing in total one hundred samples per step takes in average $32\rm ms$. The train and execution time could be further decreased by parallelizing the computation on GPUs.

In the test setup, observations of obstacles are manipulated to create scenarios with novel observations that could break the trained model. In one scenario, sensor noise is simulated by adding Gaussian noise $\sim $$N(\mu = 0, \sigma=.5)$ on the observation of position in $m$ and velocity in $\frac{m}{s}$. In another scenario, observations are randomly dropped with a probability of $20\%$. In a third and fourth scenario that simulate sensor failure, the obstacle position and velocity is masked, respectively. None of the manipulations were applied at training time.

\subsubsection{Regional novelty detection} \label{sec:reg_novel}
\begin{figure}[t]
  \centering
    \includegraphics [trim=0 0 0 0, clip, angle=0, width=1.0\columnwidth,
  keepaspectratio]{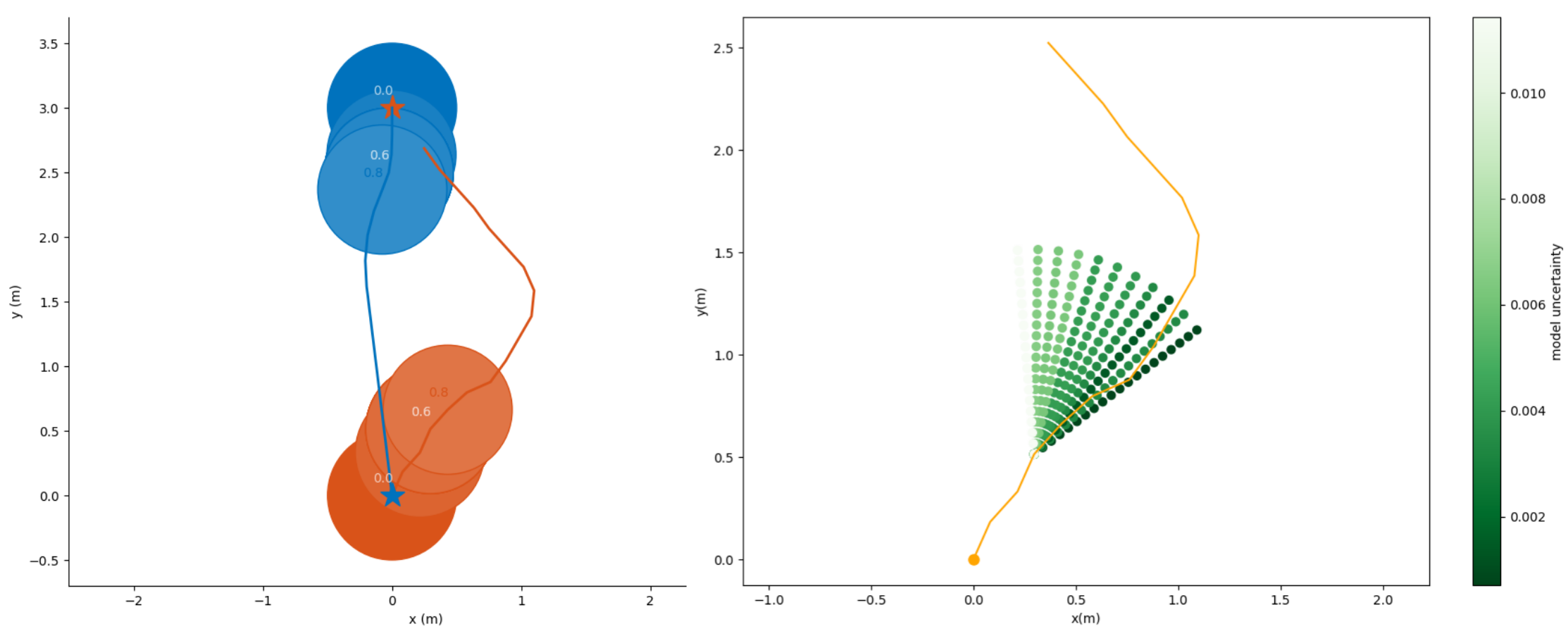}
  \caption{Regional identification of uncertainty. An uncertainty-aware agent (orange) avoids a dynamic obstacle (blue) that is observed with noise. At one time step, collision predictions for actions in the direction of the obstacle (light green lines) are assigned a higher uncertainty than for actions in free space (dark green lines). The agent selects an action with low uncertainty to cautiously avoid the obstacle.}
  \label{fig:novelty_dir} 
    \centering
  \vspace{-.0in}
\end{figure}

\begin{figure}[t]
  \centering
  \begin{subfigure}{0.45\columnwidth}
      \centering
      \includegraphics [trim=0 0 0 0, clip, width=1.0\textwidth, angle = 0]{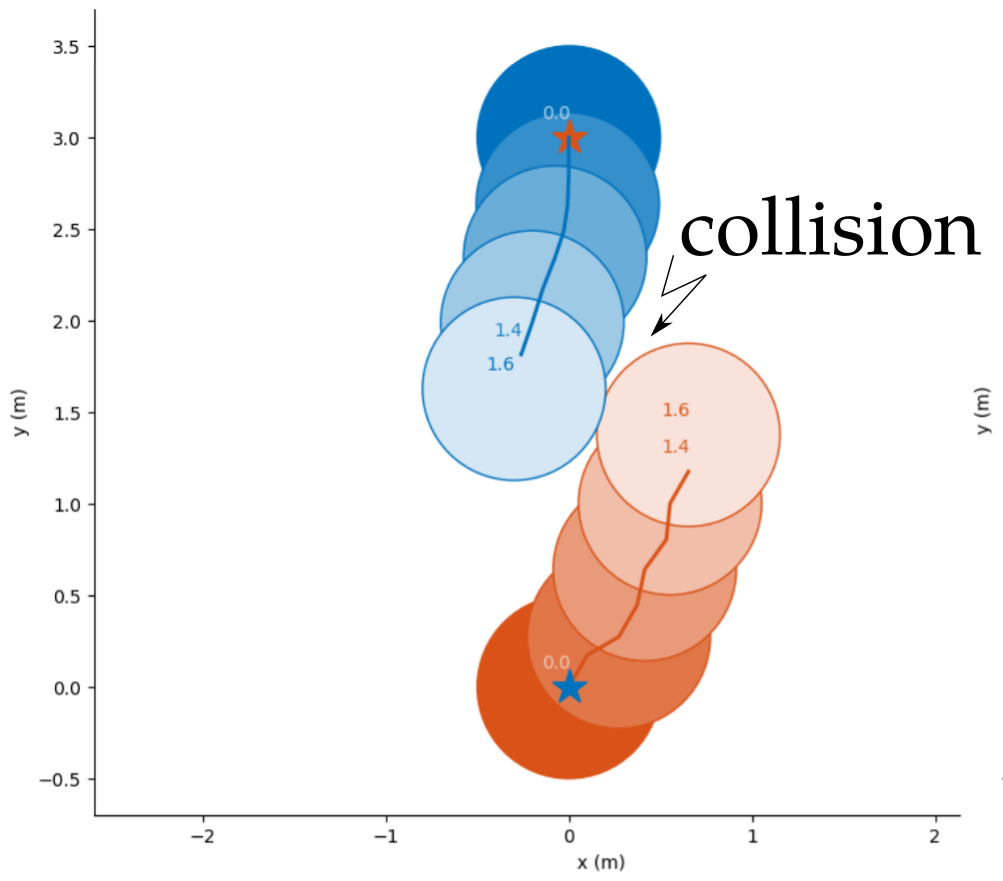}
      \caption{uncertainty-unaware}
      \label{fig:novelty_case_uu} 
  \end{subfigure}
  \begin{subfigure}{0.48\columnwidth}
      \centering
      \includegraphics [trim=0 0 0 0, clip, width=1.0\textwidth, angle = 0]{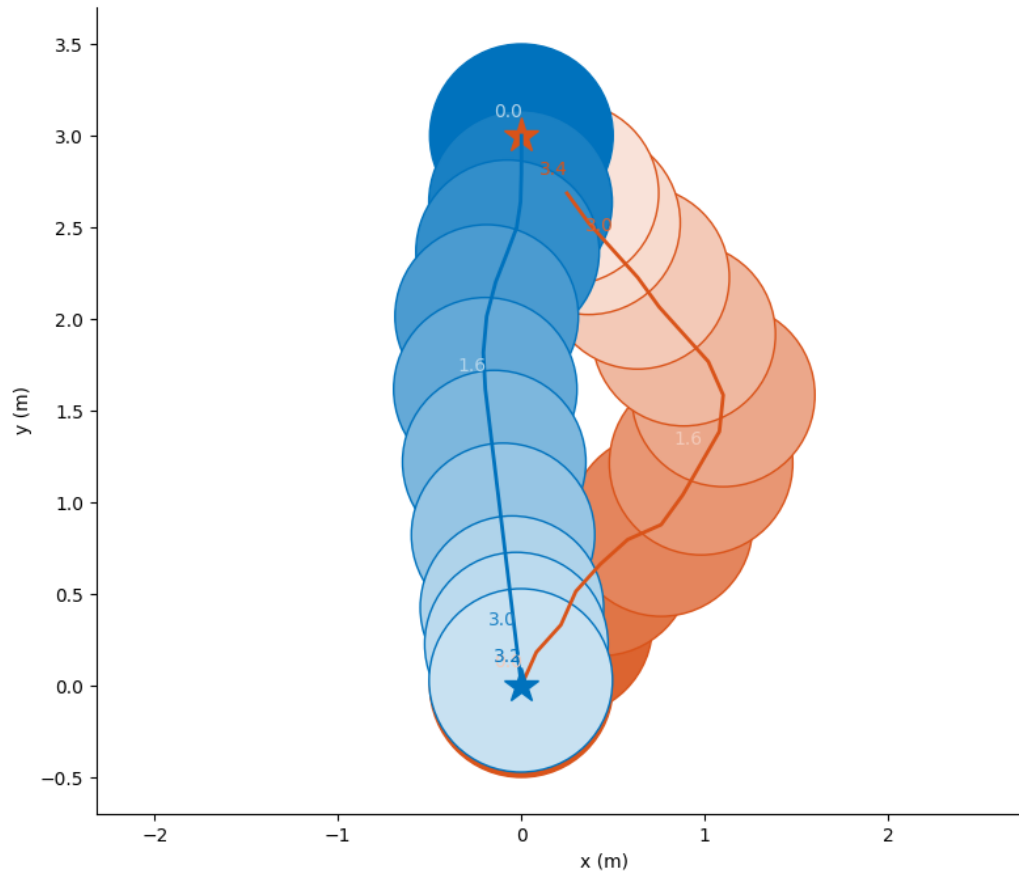}
      \caption{uncertainty-aware}
      \label{fig:novelty_case_ua} 
  \end{subfigure}
  \caption{Cautious avoidance in novel scenarios. An agent (orange) is trained to avoid dynamic RVO agents (blue) that are observed without noise. On test, Gaussian noise is added to the observation and an uncertainty-unaware model in~\cref{fig:novelty_case_uu} fails to generalize and causes a collision. The proposed uncertainty-aware agent in~\cref{fig:novelty_case_ua} acts more cautiously on novel observations and avoids the obstacle successfully.}
  \label{fig:novelty_case} 
  \centering
  \vspace{-.0in}
\end{figure}

~\Cref{fig:novelty_dir} shows that the proposed model continues to regionally identify novel obstacle observations in a higher dimensional observation space. In the displayed experiment, an uncertainty-aware agent (orange) observes a dynamic obstacle (blue) with newly added noise and evaluates actions to avoid it. The collision predictions for actions in the direction of the obstacle (light green lines) have higher uncertainty than for actions into free-space (dark green lines). The difference in the predictive uncertainties from left to right, although being stochastic and not perfectly smooth, is used by the MPC to steer the agent away from the noisy obstacle and cautiously avoid it without a collision (orange/yellow line). ~\Cref{fig:novelty_case_ua} shows the full trajectory of the uncertainty-aware agent and illustrates how an uncertainty-unaware agent in~\cref{fig:novelty_case_uu} with same speed and radius fails to generalize to the novel noise and collides with the obstacle after five time steps. 



\subsubsection{Novel scenario identification with uncertainty}
~\Cref{tab:incr_uncertainty} shows that overall model uncertainty is high in every of the tested novel scenarios, including the illustrated case of added noise. The measured uncertainty is the sum of variance of the collision predictions for each action at one time step. The uncertainty values have been averaged over $20$ sessions with random initialization, $50$ episodes and all time steps until the end of each episode. As seen in~\cref{tab:incr_uncertainty} the uncertainty in a test set of the training distribution is relatively low. All other scenarios cause higher uncertainty values and the relative magnitude of the uncertainty values can be interpreted as how novel the set of observations is for the model, in comparison to the training case. 

\begin{table*}[tp]
\centering
\vspace{0.1in}
\begin{tabular}{||c||c||c|c|c|c||}  
 \hline
  & Training & Added noise & Dropped observations & Masked vel. info. & Masked pos. info. \\ [0.5ex] 
 \hline\hline 
 $E(\rm Var(\rm P_{\rm coll}))$ & 0.363 & 0.820 & 1.93 & 1.37 & 2.41 \\[0.5ex] \hline
 $\sigma(\rm Var(\rm P_{\rm coll}))$ & 0.0330 & 0.0915 & 0.134 & 0.0693 & 0.0643 \\ \hline
\end{tabular}
\caption{Increased uncertainty in novel scenarios. In each of four novel test scenarios, the uncertainty of collision predictions $Var(\rm P_{\rm coll})$ is higher than on samples from the seen training distribution. }
\label{tab:incr_uncertainty}
\end{table*}


\begin{figure}[t]
  \centering
    \includegraphics [trim=0 0 0 0, clip, angle=0, width=1.0\columnwidth,
  keepaspectratio]{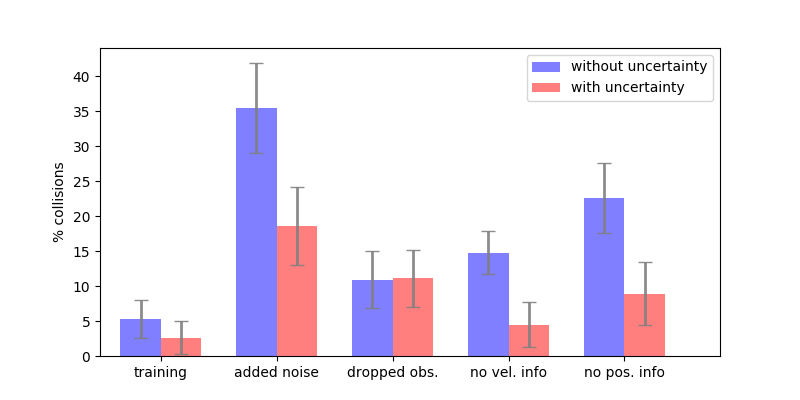}
  \caption{Fewer collisions in novel cases. The proposed uncertainty-aware model (red) causes fewer collisions than the uncertainty-unaware baseline (blue) in novel cases. Through the regional increase of uncertainty in the obstacle's direction, the model prefers actions that more cautiously avoids the obstacle than the baseline.}
  \label{fig:novelty_coll} 
    \centering
  \vspace{-.in}
\end{figure}



\subsubsection{Fewer collisions in novel scenarios}
The proposed model uses the uncertainty information to act more cautiously and be more robust to novel scenarios. ~\Cref{fig:novelty_coll} shows that this behavior causes fewer collisions during the novel scenarios than an uncertainty-unaware baseline. The proposed model (red) and the baseline (blue) perform similarly well on samples from the training distribution. In the test scenarios of added noise, masked position and masked velocity information, the proposed model causes fewer collisions and is more robust to the novel class of observations. In the case of dropped observations, both models perform similarly well, in terms of collisions, but the uncertainty-unaware model was seen to take longer to reach the goal. The baseline model has been trained with the same hyperparameters and environment except that the variance penalty $\lambda_v$ is set to zero. 

\subsubsection{Generalization to other novel scenarios}
In all demonstrated cases one could have found a model that generalizes to noise, masked position observations, etc. However, one cannot design a simulation that captures all novel scenarios that could occur in real life. A significantly novel event should be recognized with a high model uncertainty. In the pedestrian avoidance task, novel observations might be uncommon pedestrian behavior, e.g. an uncollaborative pedestrian on a personal vehicle. But really all forms of observations that are novel to the deployed model should be identified and reacted upon by driving more cautiously. The shown results suggest that model uncertainty is able to identify such observations and that the MPC selects actions with extra buffer space to avoid these pedestrians cautiously. 

\subsection{Using uncertainty to escape local minima}
\begin{figure}[t]
  \centering
  \includegraphics [trim=0 0 0 0, clip, angle=0, width=1.0\columnwidth,
  keepaspectratio]{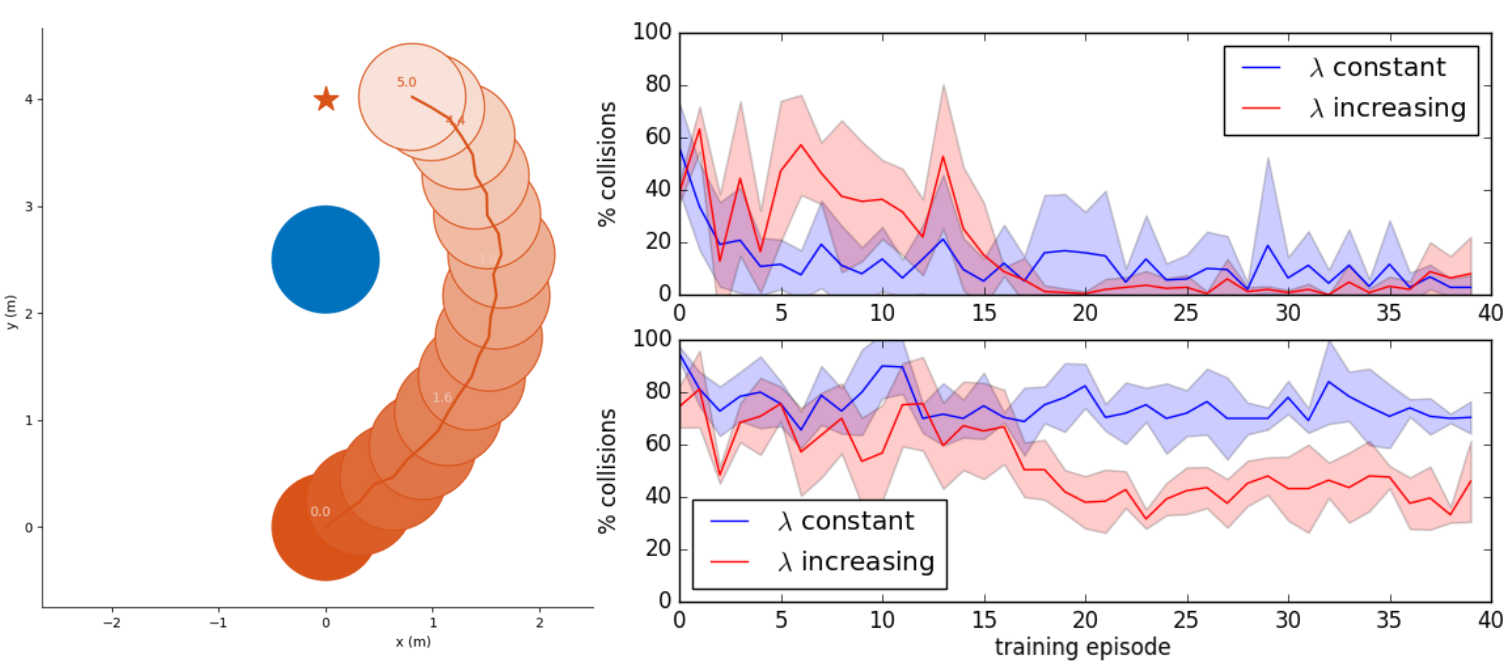}
  \caption{Escaping local minima. The training process of two policies with a constant penalty on uncertain actions $\lambda_v$(blue) and with an increasing $\lambda_v$(red) are compared. In an easy avoidance case (right-top), both policies find a good policy that leads to near-zero collisions (y-axis). In a more challenging avoidance case (right-bottom), the proposed increasing $\lambda_v$ policy, that explores in early stages, finds a better minimum than with a constant $\lambda_v$.}
  \label{fig:not_stuck} 
 \vspace{-.in}
\end{figure}

This work increases the variance penalty $\lambda_v$ to avoid getting stuck in local minima of the MPC optimization during the training process. \Cref{fig:not_stuck} shows that the proposed algorithm with increasing $\lambda_v$ can escape a local minimum by encouraging explorative actions in the early stages of training. For the experiment, an agent (orange) was trained to reach a goal (star) that is blocked by a static obstacle (blue) by continuously selecting an action (left plot). In an easy avoidance case, the obstacle is placed further away from the agent's start position (in dark orange); in a challenging case closer to the agent. A close obstacle is challenging, as the agent is initially headed into the obstacle direction and needs to explore avoiding actions. The collision estimates of the randomly initialized networks are uninformative in early training stages and the goal cost drives the agent into the obstacle. A negative variance penalty $\lambda_v$ in early stages forces the agent to explore actions away from the goal and avoid getting stuck in a local minimum. 

\Cref{fig:not_stuck} displays that, in the challenging training case, the agent with a constant $\lambda_v$ fails to explore and the algorithm gets stuck in a bad local minimum (bottom-right plot: blue), where 80\% of the runs end in a collision. The policy with an increasing $\lambda_v$, and the same hyperparameters (bottom-right plot: red), is more explorative in early stages and converges to a lower minimum in an average of five sessions. In the easy test case, both algorithms perform similarly well and converge to a policy with near-zero collisions (top-right plot).

%% file: discussion.tex
\section{Discussion and Future Work} \label{sec:discussion}
\subsection{Accurately calibrated model uncertainty estimates}
In another novel scenario, an agent was trained to avoid collaborative RVO agents and tested on uncollaborative agents. The uncertainty values did not significantly increase, which can be explained by two reasons. First, uncollaborative agents could not be seen as novel for the model; possibly, because RVO agents, further away from the agent also act in a straight line. The fact that humans think that uncollaborative agents might be novel for a model that has only been trained on collaborative agents, does not change the fact that the model might be generalizable enough to not see it as novel. Another explanation is the observed overconfidence of dropout as an uncertainty estimate. Future work will find unrevealed estimates of model uncertainty for neural networks that provide stronger guarantees on the true model uncertainty. 

%% file: conclusion.tex
\section{Conclusion} \label{sec:conclusion}
This work has developed a Safe RL framework with model uncertainty estimates to cautiously avoid dynamic obstacles in novel scenarios. An ensemble of LSTM networks was trained with dropout and bootstrapping to estimate collision probabilities and gain predictive uncertainty estimates. The magnitude of the uncertainty estimates was shown to reveal novelties in a variety of scenarios, indicating that the model \textit{knows what it does not know}. The regional uncertainty increase in the direction of novel obstacle observations is used by an MPC to act more cautious in novel scenarios. The cautious behavior made the uncertainty-aware framework more robust to novelties and safer than an uncertainty-unaware baseline. This work is another step towards opening up the vast capabilities of deep neural networks for the application in safety-critical tasks.